\RequirePackage{fix-cm}
\documentclass{article}

\usepackage{microtype}
\usepackage{graphicx}
\usepackage{subfigure}
\usepackage{booktabs} 
\usepackage{placeins}

\usepackage{amsmath,amssymb}
\usepackage{xcolor}
\usepackage{colortbl}
\definecolor{baselineRed}{RGB}{252,220,218}
\newcommand{\redmath}[1]{\begingroup\setlength{\fboxsep}{1pt}\colorbox{baselineRed}{$\displaystyle #1$}\endgroup}
\usepackage{tikz}
\usetikzlibrary{shapes.misc}

\usepackage[hypertexnames=false]{hyperref}

\usepackage[accepted]{mlsys2025}

\makeatletter
\renewcommand{\mlsys@appearing}{Preprint.}
\makeatother
\hypersetup{pdfsubject={}}

\usepackage{algorithm}
\usepackage{algorithmic}

\mlsystitlerunning{Split-KV: Context Parallelism for Training Diffusion Drafters}

\begin{document}

\twocolumn[
\mlsystitle{Split-KV: Efficient Context Parallelism for Training Diffusion Drafters}

\mlsyssetsymbol{equal}{*}

\begin{mlsysauthorlist}
\mlsysauthor{Geng Zhang}{equal,to}
\mlsysauthor{Fangzhao Zhang}{equal,to2}
\mlsysauthor{Cong Xie}{to3}
\mlsysauthor{Tiantian Fan}{to3}
\mlsysauthor{Yang You}{to}
\end{mlsysauthorlist}

\mlsysaffiliation{to}{National University of Singapore}
\mlsysaffiliation{to2}{Stanford University}
\mlsysaffiliation{to3}{ByteDance Seed}
\mlsyscorrespondingauthor{Yang You}{youy@comp.nus.edu.sg}

\mlsyskeywords{Speculative Decoding, Diffusion Language Models, Context Parallelism, Distributed Training}

\vskip 0.3in

\begin{abstract}
Diffusion drafters accelerate speculative decoding by denoising multiple draft tokens in parallel.
Their training uses anchor sampling to construct compact draft queries and performs cross-attention with a block-sparse mask over a much longer context.
We empirically show that acceptance length remains stable as context scales with a fixed query budget, while increasing the query budget yields diminishing quality gains, jointly establishing a practical regime with queries far shorter than the context.
No existing context parallelism (CP) method jointly exploits this query--context asymmetry and meets the nonlinear, sample-dependent load-balancing requirement created by anchor sampling.
We present Split-KV, an exact CP method that keeps key/value shards local and limits per-layer attention communication to query-sized tensors, making this communication independent of context length.
Its global query view further enables a cost-aware longest-processing-time-first (LPT) strategy that balances context key/value chunks by their realized attention cost.
At a CP size of 64, Split-KV retains 85--91\% weak-scaling efficiency and achieves up to $3.67\times$ speedup over the fastest available CP baseline.
LPT achieves $<$1\% imbalance and plans three orders of magnitude faster than the best load-balancing baseline.
Code is available at \url{https://github.com/zxgx/verl}.
\end{abstract}
]


\printAffiliationsAndNotice{\textsuperscript{*}Equal contribution. Work done during internship at ByteDance Seed.}%

\section{Introduction}
\label{sec:intro}

Speculative decoding accelerates large language model inference by using a lightweight drafter to propose multiple tokens and verifying them in one target-model forward pass \cite{leviathan2023specdec,chen2023specsampling}.
Diffusion drafters generate proposals by denoising a block of tokens in parallel, avoiding the sequential forward passes required by autoregressive drafters \cite{blockdiffusion2025,specdiff2024,diffuspec2025}.
Diffusion drafter training uses anchor sampling to construct compact draft queries and performs cross-attention with a block-sparse mask over a potentially much longer context.

Diffusion drafter training decouples the attention query length $Q$, determined by the anchor budget, from the context length $S$, determined by the training samples.
However, how the relationship between $Q$ and $S$ affects drafter quality is still underexplored: scaling $S$ could require proportionally more queries to preserve acceptance length, eliminating the systems benefit of this separation.
Our controlled SFT studies instead show that acceptance length remains stable as context scales with a fixed query budget, while increasing the query budget yields diminishing quality gains.
Together, these results establish a practical $Q\ll S$ regime for long-context diffusion drafter training.

Existing context parallelism (CP) methods do not fully exploit this training regime.
Ulysses, Ring Attention, and RingX retain attention communication that grows with $S$, even when $Q$ remains fixed \cite{jacobs2023ulysses,liu2023ringattention,zhu2024ringflashattention,yin2025ringx}.
LV-XAttn keeps key/value tensors stationary and makes its attention communication independent of context length, but streams query-side state through a sequential hop loop and preserves contiguous key/value shards \cite{chang2025lvxattndistributedcrossattentionlong}.
Thus, context-independent attention communication alone does not address the distinct load-balancing requirement of diffusion-drafter training.

Anchor sampling also changes how attention work is distributed.
Unlike causal self-attention, whose per-position cost follows a regular linear profile, sampled anchors produce a nonlinear and sample-dependent context-access profile.
Zigzag balances causal self-attention by pairing mirrored positions \cite{zhu2024ringflashattention,yin2025ringx}, but this position-based rule cannot capture anchor-induced context access.
Dynamic Context Parallelism (DCP) handles general attention masks through hypergraph partitioning but incurs substantial planning overhead, while RingX and LV-XAttn provide no load-balancing mechanism \cite{jiang_dcp2025,yin2025ringx,chang2025lvxattndistributedcrossattentionlong}.
Existing CP methods therefore either balance this workload inaccurately, plan it expensively, or do not balance it at all.

We introduce \emph{Split-KV} to address both limitations.
Split-KV keeps key/value shards local throughout the forward and backward passes and limits per-layer attention communication to query-sized tensors.
It exactly reconstructs the original attention while making this communication independent of context length.
The resulting global query view decouples context key/value work from query ownership, enabling a cost-aware longest-processing-time-first (LPT) strategy that assigns key/value chunks by their realized attention cost.

Our contributions are summarized as follows:
\begin{itemize}
\item We show that acceptance length remains stable as context scales with a fixed query budget, while increasing the query budget yields diminishing quality gains, jointly establishing a practical $Q\ll S$ regime.
\item We design Split-KV, an exact CP method that keeps key/value shards local and limits per-layer attention communication to query-sized tensors, making this communication independent of context length.
\item We design a cost-aware LPT strategy that uses Split-KV's global query view to balance the nonlinear, sample-dependent attention work.
\item At a CP size of 64, Split-KV retains 85--91\% weak-scaling efficiency and achieves up to $3.67\times$ speedup over the fastest available CP baseline; LPT achieves $<1\%$ imbalance and plans three orders of magnitude faster than the best load-balancing baseline, DCP.
\end{itemize}

\section{Related Work}
\label{sec:related}

\subsection{Speculative Decoding with Learnable Drafters}
Speculative decoding accelerates a target model by generating a multi-token draft at low cost and verifying it in one target-model forward pass, thereby alleviating memory-bound decoding \cite{leviathan2023specdec, chen2023specsampling}.
Extensive work studies autoregressive drafters, including the EAGLE family \cite{li2026eagle} and MTP \cite{liu2024deepseek}.
These drafters follow the target-model architecture but use only a few layers to reduce draft-generation cost.
Diffusion drafters instead denoise a block of draft tokens in each forward pass.
DFlash \cite{dflash2026} adopts a block-diffusion architecture \cite{blockdiffusion2025} and uses fewer drafter layers than DiffuSpec \cite{diffuspec2025} and SpecDiff \cite{specdiff2024, sandler2026specdiff} while retaining high acceptance length.
Domino \cite{huang2026domino} and DSpark \cite{cheng2026dspark} modify the dependency structure within diffusion blocks to improve acceptance length, but retain the same anchor-based block-diffusion training pattern targeted by Split-KV.

\subsection{Context Parallelism}
Context parallelism (CP) distributes long sequences across devices to reduce per-device computation and memory at the cost of additional communication.
Sequence parallelism distributes sequence-dependent activations across devices \cite{korthikanti2023reducing}.
Ulysses uses all-to-all communication to exchange sequence shards for attention-head shards \cite{jacobs2023ulysses}.
Although Ulysses balances attention work across devices, its CP size is limited by the model's attention-head count.
Ring Attention instead partitions the sequence dimension and circulates key/value shards during local attention computation \cite{li2023sequence, liu2023ringattention}.
Zigzag Ring Attention pairs chunks at mirrored sequence positions to balance the triangular workload induced by causal self-attention \cite{yin2025ringx, zhu2024ringflashattention}.
Dynamic Context Parallelism (DCP) uses hypergraph partitioning to balance dynamic attention workloads while accounting for communication constraints \cite{jiang_dcp2025}.
Beyond methods for self-attention, RingX\footnote{RingX provides four variants.
We use RingX1, its variant for bidirectional cross-attention, and refer to it as RingX throughout this paper.} \cite{yin2025ringx} and LV-XAttn \cite{chang2025lvxattndistributedcrossattentionlong} distribute bidirectional cross-attention by broadcasting queries in the forward pass.
These methods target BERT-like \cite{liu2019roberta} or multimodal \cite{ye2025mplug} workloads without causal query--key dependencies and do not address the load imbalance introduced by anchor sampling.

\section{Motivation}
\label{sec:background}

\subsection{Diffusion Drafter Training}

Diffusion drafters use \textbf{\emph{cross-attention with a block-sparse mask}}, a pattern distinct from the self-attention used by LLMs and autoregressive drafters.
We use DFlash \cite{dflash2026} as the concrete architecture throughout the paper.
The same attention pattern appears in subsequent diffusion drafters such as Domino \cite{huang2026domino} and DSpark \cite{cheng2026dspark}, so the Split-KV design applies to them as well.

\begin{table*}[!t]
\caption{Per-rank training cost for a diffusion drafter layer. Multi-head attention is used for brevity. $H$ is hidden size, $P$ is CP size.
Communication includes both forward and backward passes.
Peak memory counts input and output tensors and communication buffers.
\textbf{Red highlights mark baseline limitations jointly addressed by Split-KV.}}
\label{tab:cost}
\vskip 0.1in
\begin{center}
\begin{small}
\renewcommand{\arraystretch}{1.35}
\begin{tabular}{lccc}
\toprule
 & communication & peak memory & load balancing \\
\midrule
no CP ($P{=}1$) & --- & $\big(2(S{+}Q)+2Q\big)H$ & N/A \\
Ulysses & $\frac{P-1}{P}\cdot\frac{4H(\redmath{S}+2Q)}{P}$ & $\big(4(S{+}Q)+2Q\big)\frac{H}{P}$ & Head sharding \\
Ring Attention & $\frac{2(3P-2)}{P}\cdot(\redmath{S}{+}Q)H$ & $\big(4(S+Q)+2Q\big)\frac{H}{P}$ & \cellcolor{baselineRed}Zigzag \\
RingX & $\frac{P-1}{P}\cdot 2H(2\redmath{S}{+}3Q)$ & $\big(4(S+Q)+2Q\big)\frac{H}{P}$ & \cellcolor{baselineRed}None \\
LV-XAttn & $\frac{2(3P-2)}{P}\cdot QH$ & $\big(2(S+Q)+8Q\big)\frac{H}{P}$ & \cellcolor{baselineRed}None \\
\rowcolor{gray!12}
\textbf{Split-KV} & $\frac{P-1}{P}\cdot 4QH$ & $\big(2(S+Q)+2PQ\big)\frac{H}{P}$ & Cost-aware LPT \\
\bottomrule
\end{tabular}
\end{small}
\end{center}
\vskip -0.1in
\end{table*}

\begin{figure}[!t]
\centering
\definecolor{ctxfill}{RGB}{173,212,235}
\definecolor{ctxline}{RGB}{66,133,180}
\definecolor{anchfill}{RGB}{247,193,80}
\definecolor{anchline}{RGB}{196,138,22}
\definecolor{mskfill}{RGB}{154,219,143}
\definecolor{mskline}{RGB}{66,158,64}
\resizebox{\linewidth}{!}{%
\begin{tikzpicture}[
  every node/.style={font=\scriptsize\sffamily},
  cell/.style={rounded corners=0.8pt, line width=0.4pt},
  cwhite/.style={cell, draw=gray!55, fill=white},
  cctx/.style={cell, draw=ctxline, fill=ctxfill},
  canch/.style={cell, draw=anchline, fill=anchfill},
  cmsk/.style={cell, draw=mskline, fill=mskfill},
]
  \def\cs{0.46}      
  \def\inset{0.04}   
  \def\ytop{2.76}    
  \def\blkL{4.39}    

  \newcommand{\drawcell}[4]{
    \draw[#1] ({#2+#3*\cs+\inset},{\ytop-(#4+1)*\cs+\inset})
      rectangle ({#2+(#3+1)*\cs-\inset},{\ytop-#4*\cs-\inset});}

  \foreach \c in {0,...,8}{\foreach \r in {0,...,5}{\drawcell{cwhite}{0}{\c}{\r}}}
  \foreach \c in {0,1}{\foreach \r in {0,...,2}{\drawcell{cctx}{0}{\c}{\r}}}
  \foreach \r in {3,...,5}{\drawcell{cctx}{0}{5}{\r}}

  \foreach \c in {0,...,5}{\foreach \r in {0,...,5}{\drawcell{cwhite}{\blkL}{\c}{\r}}}
  \foreach \r in {0,...,2}{\drawcell{canch}{\blkL}{0}{\r}}
  \foreach \c in {1,2}{\foreach \r in {0,...,2}{\drawcell{cmsk}{\blkL}{\c}{\r}}}
  \foreach \r in {3,...,5}{\drawcell{canch}{\blkL}{3}{\r}}
  \foreach \c in {4,5}{\foreach \r in {3,...,5}{\drawcell{cmsk}{\blkL}{\c}{\r}}}

  \draw[gray!55, dashed] (2.3,-0.1) -- (2.3,\ytop+0.1);

  \foreach \r/\l in {0/$a_1$, 1/[M], 2/[M], 3/$a_2$, 4/[M], 5/[M]}{
    \node[anchor=east, inner sep=2pt] at (-0.08,{\ytop-(\r+0.5)*\cs}) {\l};}
  \node[rotate=90] at (-0.85,{\ytop-1.5*\cs}) {block 1};
  \node[rotate=90] at (-0.85,{\ytop-4.5*\cs}) {block 2};

  \node at (2.07,{\ytop+0.30}) {context keys, $S$};
  \node at ({\blkL+3*\cs},{\ytop+0.30}) {block keys, $Q$};

  \foreach \c/\l in {0/$s_1$, 2/$a_1$, 5/$s_2$, 6/$a_2$}{
    \draw[gray!55] ({(\c+0.5)*\cs},-0.02) -- ({(\c+0.5)*\cs},-0.14);
    \node[anchor=north, inner sep=1.5pt] at ({(\c+0.5)*\cs},-0.16) {\l};}
  \node[anchor=north, inner sep=1.5pt] at ({\blkL+1.5*\cs},-0.16) {block 1};
  \node[anchor=north, inner sep=1.5pt] at ({\blkL+4.5*\cs},-0.16) {block 2};

  \draw[<->, gray!80] (0.02,-0.75) -- (4.12,-0.75);
  \node[fill=white, inner sep=1pt, font=\tiny\itshape\sffamily, text=gray!130]
    at (2.07,-0.75) {$S$ grows with the context};
  \draw[<->, gray!80] ({\blkL+0.02},-0.75) -- ({\blkL+6*\cs-0.02},-0.75);
  \node[fill=white, inner sep=1pt, font=\tiny\itshape\sffamily, text=gray!130]
    at ({\blkL+3*\cs},-0.75) {$Q$ fixed by the budget};

  \begin{scope}[shift={(0.55,-1.80)}, every node/.style={font=\tiny\sffamily, anchor=west}]
    \draw[cctx] (0,0.44) rectangle (0.26,0.70);
    \node at (0.34,0.57) {context window $[s_i,a_i)$};
    \draw[canch] (3.6,0.44) rectangle (3.86,0.70);
    \node at (3.94,0.57) {anchor token (clean)};
    \draw[cmsk] (0,0) rectangle (0.26,0.26);
    \node at (0.34,0.13) {mask token};
    \draw[cwhite] (3.6,0) rectangle (3.86,0.26);
    \node at (3.94,0.13) {masked out};
  \end{scope}
\end{tikzpicture}%
}
\caption{The DFlash block-sparse mask for one micro-batch with two draft
blocks ($B{=}3$). One cell is one query--key pair. Each block's anchor row
$a_i$ and its $B{-}1$ masked rows ([M]) attend the context window $[s_i,a_i)$
of their own sample and their own block.}
\label{fig:mask}
\end{figure}
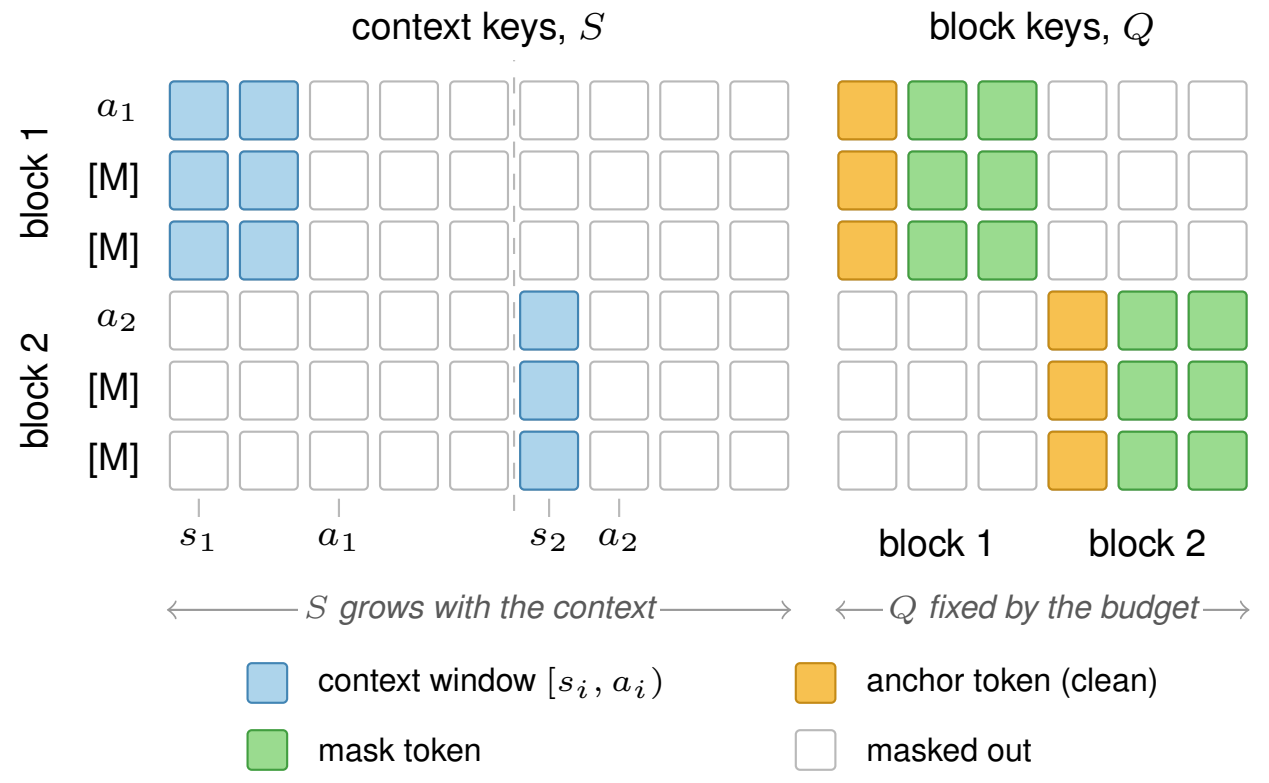

Specifically, a micro-batch of $b$ prompt--response samples is packed into $S$ context tokens $y_1,\dots,y_S$ and fed to the target model to extract intermediate hidden states.
DFlash fuses and projects these hidden states into context features $c\in\mathbb{R}^{S\times H}$ with hidden size $H$, then projects $c$ into the context keys and values at each drafter layer.
For the attention queries, anchor sampling selects $A$ response positions per sequence.
Each anchor opens a block whose remaining $B-1$ positions are mask tokens to be denoised as draft tokens.
Across the micro-batch, this process produces $N=bA$ blocks and a query length of $Q=NB$.
Anchor $i$ lies at position $a_i$ in the sequence beginning at $s_i$ and attends to its context window $\mathcal{W}_i=[s_i,a_i)$ and all $B$ tokens in its own block.
\emph{The attention is therefore causal over the context and bidirectional within each block.}
As shown in Figure~\ref{fig:mask}, FlexAttention processes all $N$ blocks in one pass with $Q$ queries and a key/value sequence comprising $S$ context tokens and $Q$ mask tokens \cite{dong2024flexattention}.
Throughout this paper, $S$ denotes the packed context length, $Q$ denotes the packed query length, and the key/value length is $S+Q$.

\subsection{Limitation 1: Communication Ignores Query-Context Asymmetry}
\label{sec:quality-context}

\begin{figure}[!t]
\centering
\includegraphics[width=\columnwidth]{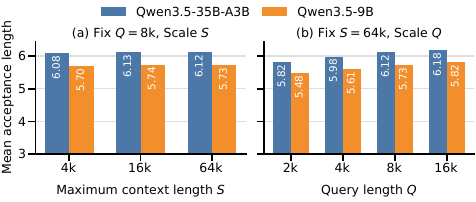}
\caption{Mean acceptance length over five downstream tasks for two controlled SFT studies.
(a) Fix $Q$--Scale $S$ varies the maximum context length at a fixed query budget.
(b) Fix $S$--Scale $Q$ varies the query budget at a fixed maximum context length.}
\label{fig:context-quality}
\end{figure}

\textbf{Query--context asymmetry }
Diffusion drafter training decouples the packed context length $S$, determined by the training samples, from the packed query length $Q$, determined by the anchor budget.
The impact of the two factors on the final drafter quality is still underexplored.
To determine whether this separation creates a useful $Q\ll S$ regime, we conduct two controlled SFT studies using DFlash drafters for Qwen3.5-9B and Qwen3.5-35B-A3B.
\emph{Fix $Q$--Scale $S$} holds the query budget fixed while increasing the maximum per-sequence context length.
\emph{Fix $S$--Scale $Q$} instead fixes the maximum context length and increases the query budget through the number of sampled anchors.
Figure~\ref{fig:context-quality} reports mean drafter acceptance length across GSM8K \cite{cobbe2021training}, MATH-500 \cite{lightman2024let}, HumanEval \cite{chen2021evaluating}, MBPP \cite{austin2021program}, and MT-Bench \cite{zheng2023judging}. 
Appendix A provides the detailed setup and results.

Figure~\ref{fig:context-quality}(a) shows that, with $Q$ fixed, scaling $S$ changes the five-task mean acceptance length by at most 0.82\% for either model.
Thus, increasing the context does not require a proportional increase in the query budget to preserve drafter quality.
Figure~\ref{fig:context-quality}(b) then fixes a long context and scales $Q$.
Acceptance length increases monotonically, confirming that query budget matters, but the gain from each doubling diminishes: the final doubling improves the mean by only 0.98\% and 1.57\% for two models separately.
The first study shows that $Q$ need not grow with $S$, while the second identifies a quality-saturation region at a query budget substantially smaller than the context.
Together, they establish our core observation: \textbf{\emph{acceptance length remains stable as context scales with a fixed query budget, while increasing the query budget yields diminishing quality gains}}, creating a practical $Q\ll S$ regime.

\textbf{Limitations of existing context parallelisms }
Table~\ref{tab:cost} shows that most existing CP methods do not exploit this query--context asymmetry.
Ulysses communicates query/key/value tensors through all-to-all collectives \cite{jacobs2023ulysses}, Ring Attention circulates key/value shards \cite{liu2023ringattention,zhu2024ringflashattention}, and RingX broadcasts key/value tensors and reduces their gradients in the backward pass \cite{yin2025ringx}.
Their total communication therefore contains an $S$-dependent term and grows with the context even when $Q$ is fixed.
LV-XAttn makes attention communication independent of $S$, but its P2P loop circulates running output and gradient state along with $Q$ and provides no mechanism for the anchor-induced load imbalance discussed next \cite{chang2025lvxattndistributedcrossattentionlong}.

\begin{figure}[!t]
\centering
\includegraphics[width=\linewidth]{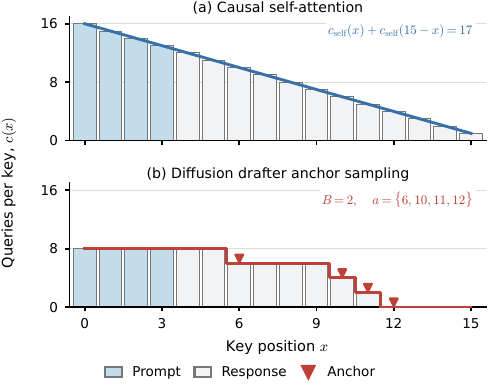}
\caption{Number of queries attending key $x$ in a 16-token sample.
Causal self-attention has a linear profile whose mirrored positions have constant total load.
Diffusion drafter anchor sampling ($B=2$, anchors $\{6,10,11,12\}$) produces a piecewise profile.}
\label{fig:profile}
\end{figure}

\subsection{Limitation 2: Inadequate Load Balancing}
\label{sec:motivation-balance}

\textbf{Nonlinear context access }
With the anchor sampling, context tokens at different positions incur different amounts of attention works, creating a unique load balance requirement different from causal self-attention.

Let the per-position load $c(x)$ be the number of query rows that attend the key at absolute context position $x$.
For causal self-attention on a sequence of length $L$, this load is linear:
\begin{equation}
\begin{aligned}
c_{\mathrm{self}}(x) &= L-x.
\end{aligned}
\label{eq:self-load}
\end{equation}
Zigzag relies on $c_{\mathrm{self}}(x)+c_{\mathrm{self}}(L-1-x)=L+1$ to achieve perfect load balance \cite{zhu2024ringflashattention}.

Diffusion drafters instead sample anchors over the response.
Each anchor contributes $B$ queries that read only the context prefix before it, so
\begin{equation}
\begin{aligned}
c_{\mathrm{anchor}}(x)
  &=B\sum_{i=0}^{N-1}\mathbf{1}\!\left[s_i\le x<a_i\right],\\
C_p&=\sum_{x\in\mathcal{K}_p}c_{\mathrm{anchor}}(x),
\end{aligned}
\label{eq:load}
\end{equation}
where $\mathcal{K}_p$ is the set of context positions assigned to rank $p$ and $C_p$ is its context-dependent attention work.
As Figure~\ref{fig:profile} shows, this load is flat over the prompt and drops only at sampled anchors in the response.
Therefore, \textit{\textbf{Anchor sampling changes the linear causal profile into a sample-dependent, nonlinear context-access profile}}.

\textbf{Limitations of existing balancing strategies }
Table~\ref{tab:cost} summarizes the balancing strategy coupled to each CP method.
Ulysses evenly shards attention heads, but its CP size $P$ is bounded by the number of heads and thus cannot scale to arbitrary CP sizes \cite{jacobs2023ulysses}.
Ring Attention adopts Zigzag \cite{zhu2024ringflashattention}, whose mirrored-position rule is \emph{inaccurate} for the nonlinear, sample-dependent profile as shown in Figure \ref{fig:profile}.
DCP can represent arbitrary sparse masks through hypergraph partitioning, but its per-step planning is \emph{expensive} (Section~\ref{sec:eval-balance}) \cite{jiang_dcp2025}.
The bidirectional cross-attention methods RingX and LV-XAttn do not provide any load-balancing mechanism and preserve contiguous context shards \cite{yin2025ringx,chang2025lvxattndistributedcrossattentionlong}.

Consequently, existing methods neither fully exploit the query--context asymmetry in their parallel design nor account for the nonlinear attention workload induced by anchor sampling in their load balancing.

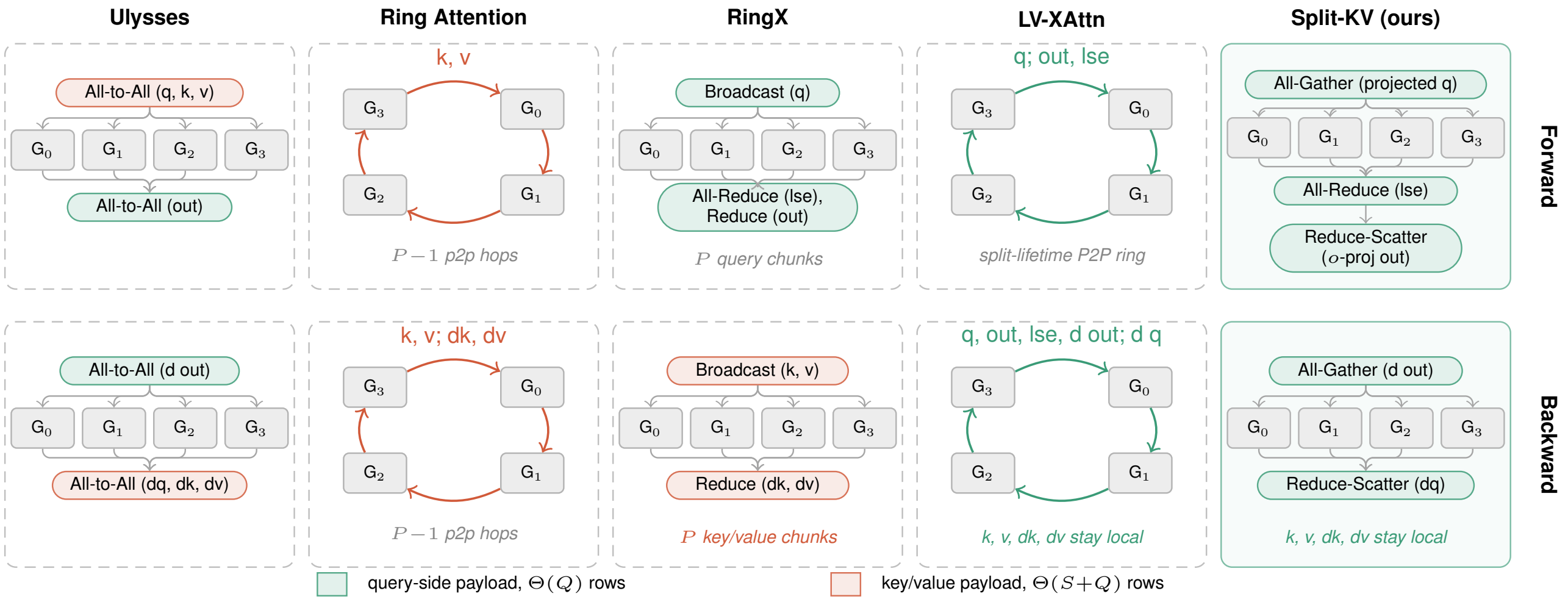
\begin{figure*}[!t]
\centering
\definecolor{ulysline}{RGB}{209,90,58}
\definecolor{skvline}{RGB}{58,155,119}
\resizebox{0.99\linewidth}{!}{%
\begin{tikzpicture}[
  y=0.78cm,
  every node/.style={font=\fontsize{4.8}{5.4}\selectfont\sffamily},
  gpu/.style={draw=gray!60, fill=gray!14, rounded corners=1.5pt, minimum width=0.60cm,
              minimum height=0.38cm, inner sep=0pt,
              font=\fontsize{4.5}{5}\selectfont\sffamily},
  kvcoll/.style={rounded rectangle, draw=ulysline!85, fill=ulysline!12, align=center,
                 inner ysep=1.5pt, inner xsep=7pt},
  qcoll/.style={rounded rectangle, draw=skvline!85, fill=skvline!14, align=center,
                inner ysep=1.5pt, inner xsep=7pt},
  flow/.style={->, gray!70, rounded corners=2pt},
  kvflow/.style={->, ulysline, semithick},
  qflow/.style={->, skvline, semithick},
  pan/.style={draw=gray!50, dash pattern=on 3pt off 2pt, rounded corners=3pt},
  ours/.style={draw=skvline!65, rounded corners=3pt, fill=skvline!4},
]
\foreach \cx in {1.5,4.4,7.3,10.2}{
  \draw[pan] (\cx-1.38,3.6) rectangle (\cx+1.38,6.6);
  \draw[pan] (\cx-1.38,0.2) rectangle (\cx+1.38,3.2);}
\draw[ours] (11.72,3.6) rectangle (14.48,6.6);
\draw[ours] (11.72,0.2) rectangle (14.48,3.2);
\foreach \cx/\name in {1.5/{Ulysses}, 4.4/{Ring Attention}, 7.3/{RingX}, 10.2/{LV-XAttn}, 13.1/{Split-KV (ours)}}{
  \node[font=\fontsize{5.8}{6.3}\selectfont\sffamily\bfseries] at (\cx,6.90) {\name};}
\node[rotate=270, font=\fontsize{5.8}{6.3}\selectfont\sffamily\bfseries] at (14.85,5.1) {Forward};
\node[rotate=270, font=\fontsize{5.8}{6.3}\selectfont\sffamily\bfseries] at (14.85,1.7) {Backward};

\node[kvcoll] (fua) at (1.5,6.00) {All-to-All (q, k, v)};
\foreach \i/\dx in {0/-1.02,1/-0.34,2/0.34,3/1.02}{
  \node[gpu] (fug\i) at (1.5+\dx,5.30) {G$_\i$};
  \draw[flow] (fua.south) -- ++(0,-0.13) -| (fug\i.north);}
\node[qcoll] (fub) at (1.5,4.60) {All-to-All (out)};
\foreach \i in {0,1,2,3}{\draw[flow] (fug\i.south) -- ++(0,-0.11) -| (fub.north);}
\node[qcoll] (bua) at (1.5,2.60) {All-to-All (d out)};
\foreach \i/\dx in {0/-1.02,1/-0.34,2/0.34,3/1.02}{
  \node[gpu] (bug\i) at (1.5+\dx,1.90) {G$_\i$};
  \draw[flow] (bua.south) -- ++(0,-0.13) -| (bug\i.north);}
\node[kvcoll, inner xsep=4pt] (bub) at (1.5,1.20) {All-to-All (dq, dk, dv)};
\foreach \i in {0,1,2,3}{\draw[flow] (bug\i.south) -- ++(0,-0.11) -| (bub.north);}

\foreach \i/\dx/\dy in {3/-0.75/5.80, 0/0.75/5.80, 1/0.75/4.75, 2/-0.75/4.75}{
  \node[gpu] (frg\i) at (4.4+\dx,\dy) {G$_\i$};}
\draw[kvflow] (frg3) to[bend left=26] node[above,font=\tiny\sffamily]{k, v} (frg0);
\draw[kvflow] (frg0) to[bend left=26] (frg1);
\draw[kvflow] (frg1) to[bend left=26] (frg2);
\draw[kvflow] (frg2) to[bend left=26] (frg3);
\node[font=\fontsize{4.8}{5.4}\selectfont\itshape\sffamily, text=gray!95] at (4.4,4.00) {$P{-}1$ p2p hops};
\foreach \i/\dx/\dy in {3/-0.75/2.40, 0/0.75/2.40, 1/0.75/1.35, 2/-0.75/1.35}{
  \node[gpu] (brg\i) at (4.4+\dx,\dy) {G$_\i$};}
\draw[kvflow] (brg3) to[bend left=26] node[above,font=\tiny\sffamily]{k, v; dk, dv} (brg0);
\draw[kvflow] (brg0) to[bend left=26] (brg1);
\draw[kvflow] (brg1) to[bend left=26] (brg2);
\draw[kvflow] (brg2) to[bend left=26] (brg3);
\node[font=\fontsize{4.8}{5.4}\selectfont\itshape\sffamily, text=gray!95] at (4.4,0.60) {$P{-}1$ p2p hops};

\node[qcoll] (fxa) at (7.3,6.00) {Broadcast (q)};
\foreach \i/\dx in {0/-1.02,1/-0.34,2/0.34,3/1.02}{
  \node[gpu] (fxg\i) at (7.3+\dx,5.30) {G$_\i$};
  \draw[flow] (fxa.south) -- ++(0,-0.13) -| (fxg\i.north);}
\node[qcoll] (fxb) at (7.3,4.60) {All-Reduce (lse),\\ Reduce (out)};
\foreach \i in {0,1,2,3}{\draw[flow] (fxg\i.south) -- ++(0,-0.11) -| (fxb.north);}
\node[font=\fontsize{4.8}{5.4}\selectfont\itshape\sffamily, text=gray!95] at (7.3,3.95) {$P$ query chunks};
\node[kvcoll] (bxa) at (7.3,2.60) {Broadcast (k, v)};
\foreach \i/\dx in {0/-1.02,1/-0.34,2/0.34,3/1.02}{
  \node[gpu] (bxg\i) at (7.3+\dx,1.90) {G$_\i$};
  \draw[flow] (bxa.south) -- ++(0,-0.13) -| (bxg\i.north);}
\node[kvcoll] (bxb) at (7.3,1.20) {Reduce (dk, dv)};
\foreach \i in {0,1,2,3}{\draw[flow] (bxg\i.south) -- ++(0,-0.11) -| (bxb.north);}
\node[font=\fontsize{4.8}{5.4}\selectfont\itshape\sffamily, text=ulysline] at (7.3,0.55) {$P$ key/value chunks};

\foreach \i/\dx/\dy in {3/-0.75/5.80, 0/0.75/5.80, 1/0.75/4.75, 2/-0.75/4.75}{
  \node[gpu] (flg\i) at (10.2+\dx,\dy) {G$_\i$};}
\draw[qflow] (flg3) to[bend left=26] node[above,font=\tiny\sffamily]{q; out, lse} (flg0);
\draw[qflow] (flg0) to[bend left=26] (flg1);
\draw[qflow] (flg1) to[bend left=26] (flg2);
\draw[qflow] (flg2) to[bend left=26] (flg3);
\node[font=\fontsize{4.8}{5.4}\selectfont\itshape\sffamily, text=gray!95] at (10.2,4.00) {split-lifetime P2P ring};
\foreach \i/\dx/\dy in {3/-0.75/2.40, 0/0.75/2.40, 1/0.75/1.35, 2/-0.75/1.35}{
  \node[gpu] (blg\i) at (10.2+\dx,\dy) {G$_\i$};}
\draw[qflow] (blg3) to[bend left=26] node[above,font=\tiny\sffamily]{q, out, lse, d out; d q} (blg0);
\draw[qflow] (blg0) to[bend left=26] (blg1);
\draw[qflow] (blg1) to[bend left=26] (blg2);
\draw[qflow] (blg2) to[bend left=26] (blg3);
\node[font=\fontsize{4.8}{5.4}\selectfont\itshape\sffamily, text=skvline] at (10.2,0.55) {k, v, dk, dv stay local};

\node[qcoll] (fsa) at (13.1,6.10) {All-Gather (projected q)};
\foreach \i/\dx in {0/-1.02,1/-0.34,2/0.34,3/1.02}{
  \node[gpu] (fsg\i) at (13.1+\dx,5.45) {G$_\i$};
  \draw[flow] (fsa.south) -- ++(0,-0.11) -| (fsg\i.north);}
\node[qcoll] (fsb) at (13.1,4.80) {All-Reduce (lse)};
\foreach \i in {0,1,2,3}{\draw[flow] (fsg\i.south) -- ++(0,-0.10) -| (fsb.north);}
\node[qcoll] (fsc) at (13.1,4.10) {Reduce-Scatter\\ ($o$-proj out)};
\draw[flow] (fsb.south) -- (fsc.north);
\node[qcoll] (bsa) at (13.1,2.60) {All-Gather (d out)};
\foreach \i/\dx in {0/-1.02,1/-0.34,2/0.34,3/1.02}{
  \node[gpu] (bsg\i) at (13.1+\dx,1.90) {G$_\i$};
  \draw[flow] (bsa.south) -- ++(0,-0.13) -| (bsg\i.north);}
\node[qcoll] (bsb) at (13.1,1.20) {Reduce-Scatter (dq)};
\foreach \i in {0,1,2,3}{\draw[flow] (bsg\i.south) -- ++(0,-0.11) -| (bsb.north);}
\node[font=\fontsize{4.8}{5.4}\selectfont\itshape\sffamily, text=skvline] at (13.1,0.55) {k, v, dk, dv stay local};

\begin{scope}[shift={(3.1,-0.15)}]
  \fill[skvline!14] (0,0) rectangle (0.28,0.28); \draw[skvline!85] (0,0) rectangle (0.28,0.28);
  \node[anchor=west] at (0.36,0.14) {query-side payload, $\Theta(Q)$ rows};
  \fill[ulysline!12] (4.9,0) rectangle (5.18,0.28); \draw[ulysline!85] (4.9,0) rectangle (5.18,0.28);
  \node[anchor=west] at (5.26,0.14) {key/value payload, $\Theta(S{+}Q)$ rows};
\end{scope}
\end{tikzpicture}%
}
\caption{Split-KV exploits $Q\ll S$ by all-gathering the global query while keeping the long key/value shards local.
The global query view decouples key/value-chunk work from query ownership and enables the cost-aware balancing design.}
\label{fig:collectives}
\end{figure*}

\section{Split-KV: Context Parallelism for Diffusion Drafters}
\label{sec:splitkv}

This section introduces Split-KV, a context-parallel method designed for diffusion-drafter training with $Q\ll S$.
Section~\ref{sec:splitkv-design} presents the parallelism design and analyzes its communication, computation and memory overhead.
Section~\ref{sec:splitkv-balance} analyzes the load skew introduced by anchor sampling and presents a low-cost longest-processing-time-first (LPT) load-balancing strategy built upon Split-KV's parallelism.

\subsection{Split-KV Parallelism Design}
\label{sec:splitkv-design}

Motivated by $Q\ll S$, Split-KV starts from the asymmetry between the $Q$ query-source rows $X\in\mathbb{R}^{Q\times H}$ and the $S+Q$ key/value-source rows $U\in\mathbb{R}^{(S+Q)\times H}$.
Split-KV partitions $X=\bigsqcup_{p=0}^{P-1}X_p$ and $U=\bigsqcup_{p=0}^{P-1}U_p$ by rows across $P$ devices, but replicates only the projected queries $q=\operatorname{QProj}(X)$ while keeping $k_p,v_p=\operatorname{KeyValueProj}(U_p)$ at rank $p$.
For one query and attention head, let $z_p=qk_p^\top/\sqrt{d}+\mathcal{M}_p$ denote the masked scores over rank $p$'s local keys, with $\mathcal{M}_p$ being the block-sparse mask and masked entries set to $-\infty$, the local attention result and its log-sum-exp (LSE) are
\begin{equation}
\begin{aligned}
\ell_p&=\log\sum_j\exp(z_{p,j}),\\
o_p&=\sum_j\exp(z_{p,j}-\ell_p)v_{p,j}
\end{aligned}
\label{eq:splitkv-local}
\end{equation}
Because the local key sets form a disjoint partition of the complete key/value sequence, the single-device result satisfies
\begin{equation}
\begin{aligned}
\ell&=\log\sum_{p=0}^{P-1}\exp(\ell_p),\\
o&=\sum_{p=0}^{P-1}\exp(\ell_p-\ell)o_p
   =\sum_{p=0}^{P-1}\bar{o}_p
\end{aligned}
\label{eq:splitkv-merge}
\end{equation}
where $\bar{o}_p=\exp(\ell_p-\ell)o_p$ is rank $p$'s additive contribution to the globally normalized output.
Writing OProj as its linear map and adding any bias once after reduction, linearity further gives
\begin{equation}
Y_p
=\left[\sum_{p=0}^{P-1}\operatorname{OProj}(\bar{o}_p)\right]_p
=\left[\operatorname{OProj}\!\left(\sum_{p=0}^{P-1}\bar{o}_p\right)\right]_p
\label{eq:splitkv-output}
\end{equation}
where $[\cdot]_p$ selects the query rows owned by rank $p$ and is implemented together with the summation by a sum reduce-scatter.
Equations~\ref{eq:splitkv-local}--\ref{eq:splitkv-output} therefore reconstruct the single-device attention exactly without moving any key/value rows; the forward pass communicates only $q$ and its projected output, and the backward pass communicates their corresponding gradients as depicted in Figure~\ref{fig:collectives}.

\begin{algorithm}[!t]
\caption{Split-KV for one draft layer on rank $p$. \textsc{AG},
\textsc{AR}, and \textsc{RS} denote all-gather, all-reduce, and
reduce-scatter.}
\label{alg:splitkv}
\small
\begin{algorithmic}[1]
\STATE \textbf{Input:} $X_p,U_p,\mathcal{M}_p$
\hfill $\triangleright$ local query rows, key/value-source rows, attention mask
\STATE \textbf{Forward pass:}
\STATE $q_p \gets \operatorname{QProj}(X_p)$
\STATE launch $q \gets \textsc{AG}(q_p)$
\STATE $k_p,v_p \gets \operatorname{KeyValueProj}(U_p)$
\hfill $\triangleright$ overlaps \textsc{AG}
\STATE wait for $q$
\STATE $(o_p,\ell_p)\gets\operatorname{Attn}(q,k_p,v_p;\mathcal{M}_p)$
\STATE $(\ell,\bar{o}_p)\gets\textsc{Merge}_{\rm AR}(o_p,\ell_p)$
\STATE $\bar{y}_p\gets\operatorname{OProj}(\bar{o}_p)$
\STATE $Y_p\gets\textsc{RS}_{\rm sum}(\bar{y}_p)$
\STATE \textbf{return} $Y_p$
\STATE \textbf{Backward pass:}
\STATE $g\gets\textsc{AG}(dY_p)$
\STATE $d\bar{o}_p\gets\operatorname{OProjBwd}(g)$
\STATE $(d o_p,d\ell_p)\gets\textsc{MergeBwd}_{\rm AR}(d\bar{o}_p)$
\STATE $(dq^{(p)},dk_p,dv_p)\gets\operatorname{AttnBwd}_{p}$
\STATE launch $dq_p\gets\textsc{RS}_{\rm sum}(dq^{(p)})$
\STATE $dU_p\gets\operatorname{KeyValueProjBwd}(dk_p,dv_p)$
\hfill $\triangleright$ overlaps \textsc{RS}
\STATE wait for $dq_p$
\STATE $dX_p\gets\operatorname{QProjBwd}(dq_p)$
\STATE \textbf{return} $dX_p,dU_p$
\end{algorithmic}
\end{algorithm}

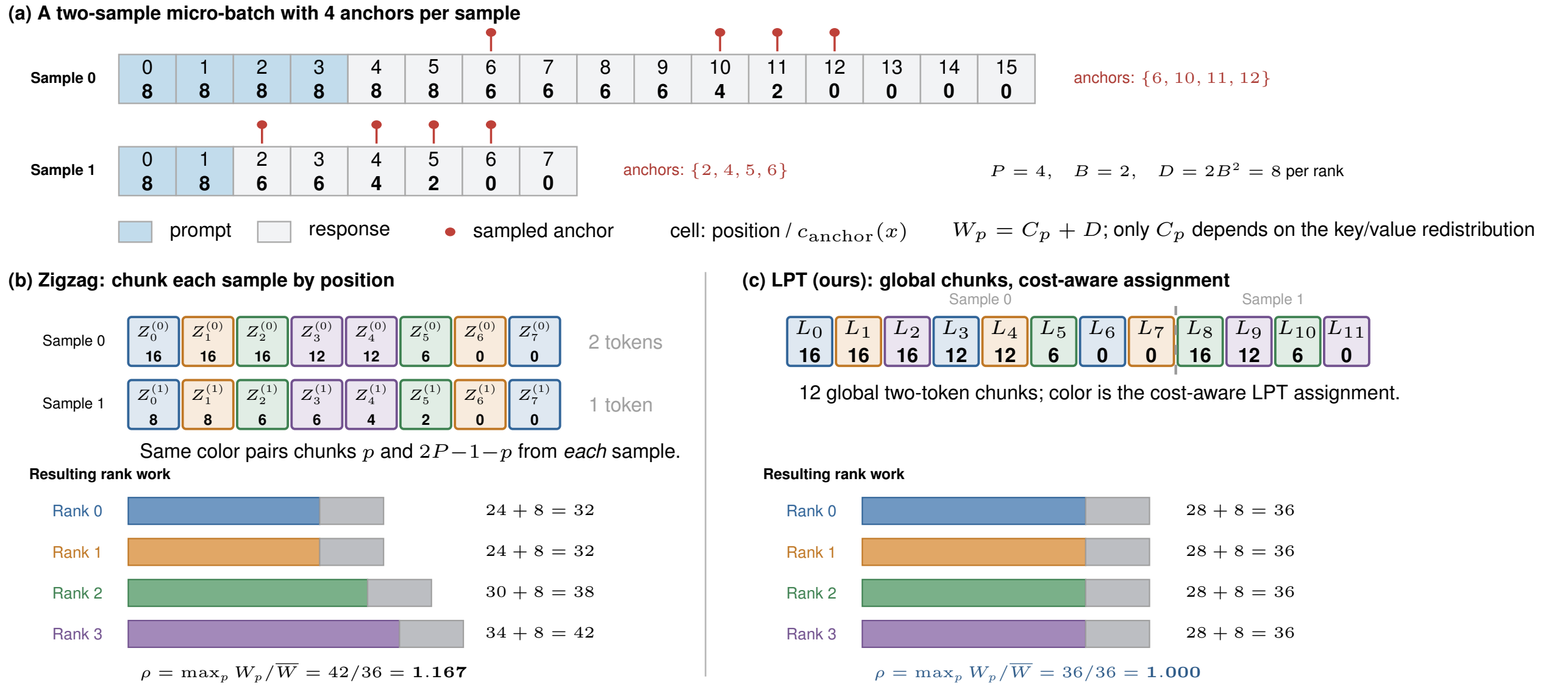
\begin{figure*}[!t]
\centering
\definecolor{figPrompt}{RGB}{194,220,236}
\definecolor{figResponse}{RGB}{242,243,245}
\definecolor{figAnchor}{RGB}{190,65,57}
\definecolor{figFixed}{RGB}{188,190,194}
\definecolor{figRankZero}{RGB}{56,112,170}
\definecolor{figRankOne}{RGB}{213,133,42}
\definecolor{figRankTwo}{RGB}{70,145,91}
\definecolor{figRankThree}{RGB}{127,91,159}
\resizebox{0.98\textwidth}{!}{%
\begin{tikzpicture}[
  y=0.84cm,
  every node/.style={font=\fontsize{4.8}{5.4}\selectfont\sffamily},
  input/.style={line width=0.45pt, draw=gray!65},
  chunk/.style={rounded corners=1pt, line width=0.7pt},
]
  \def\tw{0.61}
  \newcommand{\inputcell}[6]{
    \draw[input, fill=#6] ({#1},#2) rectangle ++(#3,0.62);
    \node[font=\tiny\sffamily, align=center] at
      ({#1+#3/2},{#2+0.31}) {\shortstack{#4\\\textbf{#5}}};
  }
  \newcommand{\costchunk}[6]{
    \draw[chunk, draw=#4!90!black, fill=#4!16]
      ({#1},#2) rectangle ++(#3,0.62);
    \node[font=\tiny\sffamily, align=center] at
      ({#1+#3/2},{#2+0.31}) {\shortstack{#5\\\textbf{#6}}};
  }
  \newcommand{\zigzagchunk}[6]{
    \draw[chunk, draw=#4!90!black, fill=#4!16]
      ({#1},#2) rectangle ++(#3,0.62);
    \node[font=\fontsize{4.5}{5}\selectfont\sffamily, align=center, inner sep=0pt] at
      ({#1+#3/2},{#2+0.31}) {\shortstack{#5\\\textbf{#6}}};
  }

  \node[anchor=west, font=\fontsize{5.8}{6.3}\selectfont\bfseries\sffamily] at (0,8.25)
    {(a) A two-sample micro-batch with 4 anchors per sample};
  \node[anchor=east, font=\fontsize{4.5}{5}\selectfont\bfseries\sffamily] at (1.18,7.43) {Sample 0};
  \foreach \position/\costval in {0/8,1/8,2/8,3/8} {
    \inputcell{1.30+\position*\tw}{7.12}{\tw}
      {\position}{\costval}{figPrompt}
  }
  \foreach \position/\costval in {
      4/8,5/8,6/6,7/6,8/6,9/6,10/4,11/2,12/0,13/0,14/0,15/0} {
    \inputcell{1.30+\position*\tw}{7.12}{\tw}
      {\position}{\costval}{figResponse}
  }
  \foreach \position in {6,10,11,12} {
    \draw[figAnchor, line width=0.65pt]
      ({1.30+(\position+0.5)*\tw},7.78) -- ++(0,0.20);
    \fill[figAnchor]
      ({1.30+(\position+0.5)*\tw},8.02) circle (0.055);
  }
  \node[anchor=west, text=figAnchor!90!black] at (11.35,7.43)
    {anchors: $\{6,10,11,12\}$};

  \node[anchor=east, font=\fontsize{4.5}{5}\selectfont\bfseries\sffamily] at (1.18,6.26) {Sample 1};
  \foreach \position/\costval in {0/8,1/8} {
    \inputcell{1.30+\position*\tw}{5.95}{\tw}
      {\position}{\costval}{figPrompt}
  }
  \foreach \position/\costval in {2/6,3/6,4/4,5/2,6/0,7/0} {
    \inputcell{1.30+\position*\tw}{5.95}{\tw}
      {\position}{\costval}{figResponse}
  }
  \foreach \position in {2,4,5,6} {
    \draw[figAnchor, line width=0.65pt]
      ({1.30+(\position+0.5)*\tw},6.61) -- ++(0,0.20);
    \fill[figAnchor]
      ({1.30+(\position+0.5)*\tw},6.85) circle (0.055);
  }
  \node[anchor=west, text=figAnchor!90!black] at (6.55,6.26)
    {anchors: $\{2,4,5,6\}$};
  \node[anchor=west] at (10.45,6.26)
    {$P=4,\quad B=2,\quad D=2B^2=8$ per rank};

  \draw[input, fill=figPrompt] (1.30,5.36) rectangle ++(0.35,0.26);
  \node[anchor=west, font=\tiny\sffamily] at (1.72,5.49) {prompt};
  \draw[input, fill=figResponse] (2.78,5.36) rectangle ++(0.35,0.26);
  \node[anchor=west, font=\tiny\sffamily] at (3.20,5.49) {response};
  \fill[figAnchor] (4.83,5.49) circle (0.055);
  \node[anchor=west, font=\tiny\sffamily] at (4.95,5.49) {sampled anchor};
  \node[anchor=west, font=\tiny\sffamily] at (7.05,5.49)
    {cell: position / $c_{\mathrm{anchor}}(x)$};
  \node[anchor=west, font=\tiny\sffamily] at (10.05,5.49)
    {$W_p=C_p+D$; only $C_p$ depends on the key/value redistribution};

  \draw[gray!55, line width=0.5pt] (7.55,-0.15) -- (7.55,4.98);

  \begin{scope}[shift={(0,0)}]
    \node[anchor=west, font=\fontsize{5.8}{6.3}\selectfont\bfseries\sffamily] at (0,4.86)
      {(b) Zigzag: chunk each sample by position};
    \node[anchor=east, font=\fontsize{4.5}{5}\selectfont\sffamily] at (1.28,4.10) {Sample 0};
    \foreach \chunkid/\costval/\rankcolor in {
        0/16/figRankZero,1/16/figRankOne,2/16/figRankTwo,
        3/12/figRankThree,4/12/figRankThree,5/6/figRankTwo,
        6/0/figRankOne,7/0/figRankZero} {
      \zigzagchunk{1.40+\chunkid*0.58}{3.79}{0.54}{\rankcolor}
        {$Z_{\chunkid}^{(0)}$}{\costval}
    }
    \node[anchor=west, font=\tiny\sffamily, text=gray!75] at (6.18,4.10)
      {2 tokens};

    \node[anchor=east, font=\fontsize{4.5}{5}\selectfont\sffamily] at (1.28,3.30) {Sample 1};
    \foreach \chunkid/\costval/\rankcolor in {
        0/8/figRankZero,1/8/figRankOne,2/6/figRankTwo,
        3/6/figRankThree,4/4/figRankThree,5/2/figRankTwo,
        6/0/figRankOne,7/0/figRankZero} {
      \zigzagchunk{1.40+\chunkid*0.58}{2.99}{0.54}{\rankcolor}
        {$Z_{\chunkid}^{(1)}$}{\costval}
    }
    \node[anchor=west, font=\tiny\sffamily, text=gray!75] at (6.18,3.30)
      {1 token};
    \node[anchor=west, font=\tiny\sffamily] at (1.40,2.69)
      {Same color pairs chunks $p$ and $2P{-}1{-}p$ from \emph{each} sample.};

    \node[anchor=west, font=\fontsize{4.5}{5}\selectfont\bfseries\sffamily] at (0.22,2.40)
      {Resulting rank work};
    \foreach \rankid/\ypos/\contextload/\fullload/\rankcolor in {
        0/1.78/24/32/figRankZero,1/1.26/24/32/figRankOne,
        2/0.74/30/38/figRankTwo,3/0.22/34/42/figRankThree} {
      \node[anchor=east, text=\rankcolor!90!black]
        at (1.25,{\ypos+0.17}) {Rank \rankid};
      \draw[draw=\rankcolor!90!black, fill=\rankcolor!72]
        (1.40,\ypos) rectangle ++({\contextload*0.085},0.34);
      \draw[draw=gray!75, fill=figFixed]
        ({1.40+\contextload*0.085},\ypos) rectangle ++(0.68,0.34);
      \node[anchor=west] at (5.08,{\ypos+0.17})
        {$\contextload+8=\fullload$};
    }
    \node[anchor=west] at (1.40,-0.10)
      {$\rho=\max_p W_p/\overline{W}=42/36=\mathbf{1.167}$};
  \end{scope}

  \begin{scope}[shift={(7.82,0)}]
    \node[anchor=west, font=\fontsize{5.8}{6.3}\selectfont\bfseries\sffamily] at (0,4.86)
      {(c) LPT (ours): global chunks, cost-aware assignment};
    \node[font=\fontsize{4.5}{5}\selectfont\sffamily, text=gray!75] at (2.66,4.62) {Sample 0};
    \node[font=\fontsize{4.5}{5}\selectfont\sffamily, text=gray!75] at (5.78,4.62) {Sample 1};
    \foreach \chunkid/\costval/\rankcolor in {
        0/16/figRankZero,1/16/figRankOne,2/16/figRankThree,
        3/12/figRankZero,4/12/figRankOne,5/6/figRankTwo,
        6/0/figRankZero,7/0/figRankOne,8/16/figRankTwo,
        9/12/figRankThree,10/6/figRankTwo,11/0/figRankThree} {
      \costchunk{0.60+\chunkid*0.52}{3.79}{0.48}{\rankcolor}
        {$L_{\chunkid}$}{\costval}
    }
    \draw[gray!80, dashed, line width=0.8pt]
      ({0.60+8*0.52-0.02},3.68) -- ({0.60+8*0.52-0.02},4.55);
    \node[anchor=west, font=\tiny\sffamily] at (0.60,3.43)
      {12 global two-token chunks; color is the cost-aware LPT assignment.};

    \node[anchor=west, font=\fontsize{4.5}{5}\selectfont\bfseries\sffamily] at (0.22,2.40)
      {Resulting rank work};
    \foreach \rankid/\ypos/\rankcolor in {
        0/1.78/figRankZero,1/1.26/figRankOne,
        2/0.74/figRankTwo,3/0.22/figRankThree} {
      \node[anchor=east, text=\rankcolor!90!black]
        at (1.25,{\ypos+0.17}) {Rank \rankid};
      \draw[draw=\rankcolor!90!black, fill=\rankcolor!72]
        (1.40,\ypos) rectangle ++(2.38,0.34);
      \draw[draw=gray!75, fill=figFixed]
        (3.78,\ypos) rectangle ++(0.68,0.34);
      \node[anchor=west] at (4.72,{\ypos+0.17})
        {$28+8=36$};
    }
    \node[anchor=west, text=figRankZero!80!black] at (1.40,-0.10)
      {$\rho=\max_p W_p/\overline{W}=36/36=\mathbf{1.000}$};
  \end{scope}
\end{tikzpicture}%
}
\caption{Load balancing of the context keys in Split-KV for the same two
packed samples and sampled anchors. A cell shows its position and realized
context cost $c_{\mathrm{anchor}}(x)$; chunk cells show the chunk label above
its summed context cost. Zigzag independently cuts each sample into $2P$
chunks and pairs mirrored positions, yielding
$(C_0,\ldots,C_3)=[24,24,30,34]$. LPT cuts the global packed context into
$PJ$ chunks and assigns them by realized chunk cost, yielding
$(C_0,\ldots,C_3)=[28,28,28,28]$. Adding the identical $D=2B^2=8$ gives
$(W_0,\ldots,W_3)=[32,32,38,42]$ and
$(W_0,\ldots,W_3)=[36,36,36,36]$, respectively.}
\label{fig:lpt-example}
\end{figure*}

RingX \cite{yin2025ringx} and LV-XAttn \cite{chang2025lvxattndistributedcrossattentionlong} expose one query shard at each communication step.
Under the anchor sampling, a key/value chunk's cost varies with the active query shard, so balancing aggregate work does not balance every synchronized communication step.
Split-KV instead all-gathers the complete $q$ before local attention, making every rank evaluate the full realized query workload against its local key/value chunks at once.
This collapses the step-dependent workload into one destination-independent cost per key/value chunk, enabling the cost-aware assignment in Section~\ref{sec:splitkv-balance}, while $Q\ll S$ keeps the additional query memory small.

Algorithm~\ref{alg:splitkv} instantiates this decomposition on rank $p$, where $X_p$, $U_p$, and $\mathcal{M}_p$ are the local query rows, the local key/value-source rows comprising the context shard and owned block rows, and the corresponding block-sparse mask.
In the forward pass, the query all-gather is overlapped with the key/value projection (lines~4--6), after which lines~7--10 evaluate Equations~\ref{eq:splitkv-local}--\ref{eq:splitkv-output}.
The backward pass applies the dual schedule (lines~13--21), overlapping the query-gradient reduce-scatter with the key/value-projection backward computation while keeping $dk_p$ and $dv_p$ local.

Table~\ref{tab:cost} summarizes the communication and peak-memory costs of Split-KV.
The forward-pass all-gather and reduce-scatter each move $\frac{P-1}{P}QH$ elements per rank, and their backward-pass duals move the same amount.
Thus Split-KV communicates $\frac{P-1}{P}4QH$ elements per layer.
For a fixed micro-batch size and anchor budget, $Q$ remains fixed as $S$ grows, so this traffic is independent of the context length; the LSE statistics contribute only a lower-order term.
For block $i$, its $B$ queries attend to $a_i-s_i$ context rows and all $B$ rows in the same block, producing $B(a_i-s_i)+B^2$ unmasked query--key pairs.
The complete block-sparse attention therefore evaluates $T_{\mathrm{attn}}=B\sum_{i=0}^{N-1}(a_i-s_i)+NB^2$ pairs, and balanced Split-KV assigns an average attention cost of $T_{\mathrm{attn}}H/P$ to each rank.
At the forward-pass peak, each rank simultaneously holds a local key/value shard of length $(S+Q)/P$, the all-gathered queries, and the partial projected outputs, giving a peak of $2(S+Q)H/P+2QH$.
The backward pass has the same peak memory.
Having established context-independent communication and destination-independent chunk costs, we now turn to assigning those chunks across ranks.

\subsection{Cost-Aware LPT Load Balancing for Split-KV}
\label{sec:splitkv-balance}

\paragraph{Cost-aware LPT}
Using the global queries exposed by Split-KV, we address the nonlinear workload identified in Section~\ref{sec:motivation-balance} with a cost-aware heuristic based on longest processing time (LPT) first strategy.
Our cost-aware LPT partitions the packed context globally into $PJ$ equal chunks of $m=S/(PJ)$ rows, where each rank must receive $J$ chunks.
This one-dimensional assignment relies on Split-KV's global query view: every rank evaluates the complete query set against whichever context chunks it owns.
The aggregate cost of each key/value chunk is therefore independent of its destination rank and can be represented by one scalar, whereas hop-based methods expose a different cost for each active query shard and require a joint assignment over chunks and communication steps.
Because every rank has the global anchor metadata $\mathcal{A}=\{(s_i,a_i)\}_{i=0}^{N-1}$, it can independently price chunk $j$, whose left boundary is $\lambda_j$, as
\begin{equation}
\hat{c}_j \;=\; B \sum_{i=0}^{N-1}
\big(\min(a_i, \lambda_j + m) - \max(s_i, \lambda_j)\big)^{+},
\label{eq:chunk}
\end{equation}
where $(\cdot)^+$ denotes the positive part.
We sort chunks by $\hat{c}_j$, the context work of chunk $j$, and repeatedly assign the remaining chunk with the longest estimated processing time (the highest cost) to the least-loaded rank with fewer than $J$ chunks.
A final locality matching maps the resulting bundles to physical ranks without changing their loads.

The plan itself requires no communication; one all-to-all then redistributes the context-source rows according to the assignment once per training step, before any draft layer executes.
This step-level key/value redistribution is separate from the per-layer communication reported in Table~\ref{tab:cost} and is amortized across all draft layers.
The draft blocks in Figure~\ref{fig:mask} remain local.
With balanced block ownership they add the same $D=(N/P)B^2$ work to every rank.
Thus rank $p$ has total work $W_p=C_p+D$.
We define the load-imbalance ratio as $\rho=\max_p W_p/\overline{W}$, where $\overline{W}=P^{-1}\sum_p W_p$.

\paragraph{A concrete example}
Figure~\ref{fig:lpt-example} compares Zigzag, the position-based strategy discussed in Section~\ref{sec:motivation-balance}, with cost-aware LPT on one realized micro-batch.
Panel (a) shows two packed samples with different lengths and anchor locations, which induce the nonlinear per-position costs from Equation~\ref{eq:load}; here $u\in\{0,1\}$ indexes the sample.
Panel (b) applies Zigzag independently to each sample, where $Z_j^{(u)}$ denotes chunk $j$, and pairs mirrored positions without considering their realized costs.
Although every rank receives the same number of chunks, their full loads become $[32,32,38,42]$.
Panel (c) instead forms global chunks $L_j$, prices them with Equation~\ref{eq:chunk}, and assigns them by LPT, producing $[36,36,36,36]$.
Because the block-local contribution $D=8$ is identical across ranks, equalizing the context contribution $C_p$ reduces $\rho$ from $1.167$ to $1.000$.

\subsection{Implementation Details}

Our implementation is based on VeOmni, a PyTorch-native backend that composes FSDP2 \cite{zhao2023pytorch} with sequence parallelism and expert parallelism on multidimensional device meshes \cite{ma2025veomni}.
Our implementation comprises approximately 4.3K lines of Python for constructing the drafter, capturing target-model hidden states, sampling anchors, building block-sparse masks, balancing the key/value chunks, computing the denoising loss and metrics, and integrating optimizer and checkpointing.

Beyond supporting diffusion-drafter training and implementing Split-KV, efficient long-context execution requires an additional system optimization.
Packed, padding-free batches reduce computation but expose variable local key/value lengths that trigger repeated FlexAttention compilation.
We therefore round each local key/value length up to a multiple of 4096, forming \textit{static-shape buckets} whose compiled kernels can be reused.
Padding is masked out before attention and loss computation.

\section{Evaluation}
\label{sec:eval}

\subsection{Experiment Setup}
\label{sec:eval-setup}

\textbf{Testbed }
We conduct all experiments on up to eight nodes, each equipped with eight accelerators.
Within a node, each accelerator has 600~GB/s of aggregate bidirectional intra-node bandwidth; across nodes, communication goes through RDMA with 400~Gb/s of inter-node bandwidth.

\textbf{Workload }
We evaluate DFlash drafters for Qwen3.5-35B-A3B and Qwen3.5-27B.
The target models are from the official Qwen3.5 model collection\footnote{https://huggingface.co/collections/Qwen/qwen35}, and the drafter model architectures are summarized in Table~\ref{tab:drafter-config} following the original DFlash configurations\footnote{https://huggingface.co/collections/z-lab/dflash} \cite{dflash2026}.

\begin{table}[t]
\centering
\caption{Architectures of the evaluated DFlash drafters, where $n_q$, $n_{kv}$, and $d_h$ denote the numbers of query heads, key/value heads, and the head dimension, respectively.}
\label{tab:drafter-config}
\small
\setlength{\tabcolsep}{2.5pt}
\resizebox{\columnwidth}{!}{%
\begin{tabular}{lrrrrrr}
\toprule
Target & Layers & $d_{\mathrm{model}}$ & $d_{\mathrm{ffn}}$ & $n_q$ & $n_{kv}$ & $d_h$ \\
\midrule
Qwen3.5-35B-A3B & 8 & 2048 & 6144 & 32 & 4 & 128 \\
Qwen3.5-27B & 5 & 5120 & 17408 & 32 & 8 & 128 \\
\bottomrule
\end{tabular}
}
\end{table}

\textbf{Evaluation protocol } 
Because diffusion drafters are typically trained for multiple epochs \cite{dflash2026,huang2026domino,cheng2026dspark}, we follow the offline training setting to directly feed target-model hidden states to the drafter and measure only drafter training step time.
This offline setting excludes target-model feature extraction time.
To eliminate variable-length sequences as a confounding factor, we use randomly generated fixed-length samples with a $1{:}3$ prompt-to-response ratio.
Following DFlash, we set the diffusion block size to $B=16$ and vary the anchor budget $A$ to control the query length.
We report per-sample context and query lengths when describing each benchmark configuration.
For micro-batch size $b$, the packed $S$ and $Q$ defined in Section~\ref{sec:background} are obtained by multiplying the corresponding per-sample lengths by $b$.

To isolate context parallelism, we set the DP size to one and world size equal to CP size $P\in\{8,16,32,64\}$.
We report end-to-end drafter training step time averaged over three steps after five warmup steps.

\begin{figure}[t]
\centering
\includegraphics[width=\columnwidth]{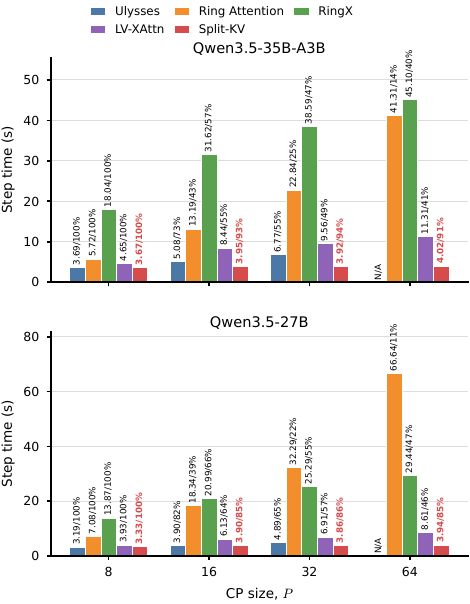}
\caption{Drafter training step time under weak scaling.
Each bar is labeled with absolute time in seconds and weak-scaling efficiency $E_{\mathrm{weak}}(P)=T_8/T_P$, separated by a slash.
Lower time and higher efficiency are better; 100\% efficiency denotes unchanged drafter training step time as the context and $P$ grow proportionally.
N/A denotes an unsupported configuration.}
\label{fig:weak-scaling}
\end{figure}

\textbf{Baselines }
To evaluate CP methods, we compare Split-KV with four representative baselines.
\textbf{Ulysses} \cite{jacobs2023ulysses} shards attention heads through all-to-all communication.
\textbf{Ring Attention} \cite{liu2023ringattention,zhu2024ringflashattention} rotates key/value shards with a zigzag-balanced schedule.
\textbf{RingX} \cite{yin2025ringx} moves queries in the forward pass, then broadcasts and reduces key/value gradients in the backward pass.
\textbf{LV-XAttn} \cite{chang2025lvxattndistributedcrossattentionlong} circulates query-side attention state through P2P rings.

To fairly evaluate load balancing, all baselines use Split-KV and differ only in how context key/value chunks are assigned to ranks.
\textbf{No balance} preserves the contiguous chunk assignment.
\textbf{Zigzag} applies a mirrored chunk order \cite{zhu2024ringflashattention}.
\textbf{DCP} invokes KaHyPar for hypergraph partitioning \cite{jiang_dcp2025,schlag2016kahypar}.
\textbf{LPT} is our cost-aware strategy.

\subsection{Weak Scaling of Drafter Training Step Time}
\label{sec:eval-weak}

Following a weak-scaling configuration, we increase $P$ from 8 to 64 and scale the per-sample context length proportionally from $128\mathrm{k}$ to $1\mathrm{M}$ while fixing the per-sample query length at $8\mathrm{k}$ and $b=8$.
Consequently, the packed context length $S$ grows from $1\mathrm{M}$ to $8\mathrm{M}$, while the global packed query length remains $Q=64\mathrm{k}$.
The local context shard remains $S/P=128\mathrm{k}$ rows per rank.
Figure~\ref{fig:weak-scaling} reports the drafter training step time of every method at each scale.

Split-KV has the lowest reported drafter training step time in six of the eight model-scale configurations, ties Ulysses in a seventh, and provides $1.01\times$--$2.81\times$ speedup where it is strictly faster.
Across the $8\times$ scale-out, its drafter training step time remains within 3.67--4.02~s for Qwen3.5-35B-A3B and 3.33--3.94~s for Qwen3.5-27B.
At $P=64$, these results correspond to 91\% and 85\% weak-scaling efficiency, respectively, while Split-KV is $2.81\times$ and $2.19\times$ faster than the fastest available baseline.
The near-flat times reflect that its attention communication is independent of context length; the remaining efficiency loss comes from collective latency and synchronization as $P$ increases, even though the query payload stays fixed.

Ring Attention and RingX degrade rapidly because their training communication still carries context-sized tensors, whose size grows with $S$ in this experiment.
Ulysses remains competitive at small scales but is unsupported at $P=64$ because of its head-sharding constraint.
LV-XAttn's attention communication is also independent of context length, but its sequential hop loop and contiguous, unbalanced key/value placement keep it slower than Split-KV.

\subsection{Strong Scaling of Drafter Training Step Time}
\label{sec:eval-strong}

\begin{figure}[!t]
\centering
\includegraphics[width=\columnwidth]{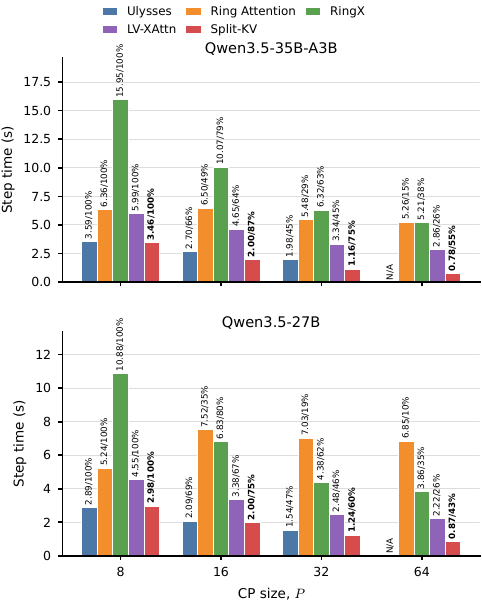}
\caption{Drafter training step time under strong scaling.
Each bar is labeled with absolute time in seconds and strong-scaling efficiency $E_{\mathrm{strong}}(P)=(T_8/T_P)/(P/8)$, separated by a slash.
Lower time and higher efficiency are better; the gap from 100\% quantifies the scale-out loss.
N/A denotes an unsupported configuration.}
\label{fig:strong-scaling}
\end{figure}

Following a strong-scaling configuration, we fix the total workload per training step at $b=4$, with a per-sample context length of $256\mathrm{k}$ and a per-sample query length of $8\mathrm{k}$, while increasing $P$ from 8 to 64.
The packed lengths are therefore $S=1\mathrm{M}$ and $Q=32\mathrm{k}$, and the local context shard $S/P$ decreases from $128\mathrm{k}$ to $16\mathrm{k}$ rows per rank.
Figure~\ref{fig:strong-scaling} reports the drafter training step time of every method as the fixed workload is distributed over more ranks.

Split-KV has the lowest reported drafter training step time in seven of the eight model-scale configurations and provides $1.04\times$--$3.67\times$ speedup over the fastest available baseline in those settings.
It reduces drafter training step time from 3.46 to 0.78~s for Qwen3.5-35B-A3B and from 2.98 to 0.87~s for Qwen3.5-27B, corresponding to $4.44\times$ and $3.43\times$ speedups under an $8\times$ increase in parallelism.
At $P=64$, Split-KV therefore retains 55\% and 43\% strong-scaling efficiency, respectively.
The efficiency gap arises because local attention work shrinks with $S/P$, whereas the global query payload remains fixed and collective latency and synchronization grow with $P$.

Ulysses is comparable to Split-KV at the smallest scale for Qwen3.5-27B but is unsupported at $P=64$.
Ring Attention and RingX retain context-dependent traffic, while LV-XAttn avoids that traffic but executes a sequential query-state hop loop without balancing key/value work.
Consequently, even though LV-XAttn is the fastest available baseline at $P=64$, Split-KV remains $3.67\times$ and $2.55\times$ faster for Qwen3.5-35B-A3B and Qwen3.5-27B, respectively.

\subsection{Query-to-Context Ratio Ablation}
\label{sec:eval-qkv}

We evaluate query-to-context sensitivity on a fixed four-node cluster with $P=32$ (32 accelerators).
Across all configurations, we hold the per-sample context length at $S_{\mathrm{sample}}=128\mathrm{k}$ and the packed query length at $Q=64\mathrm{k}$.
We vary the per-sample query length $Q_{\mathrm{sample}}$ together with the micro-batch size $b$ using $(Q_{\mathrm{sample}},b)=(32\mathrm{k},2)$, $(16\mathrm{k},4)$, $(8\mathrm{k},8)$, and $(4\mathrm{k},16)$, so that $Q=bQ_{\mathrm{sample}}$ remains fixed.
The packed context length $S=bS_{\mathrm{sample}}$ is therefore $256\mathrm{k}$, $512\mathrm{k}$, $1\mathrm{M}$, and $2\mathrm{M}$, giving $Q{:}S$ ratios of $1{:}4$, $1{:}8$, $1{:}16$, and $1{:}32$.
Because each query attends only within its own sample, the nominal query--context attention work $Q S_{\mathrm{sample}}$ also remains fixed.

\begin{figure}[!t]
\centering
\includegraphics[width=\columnwidth]{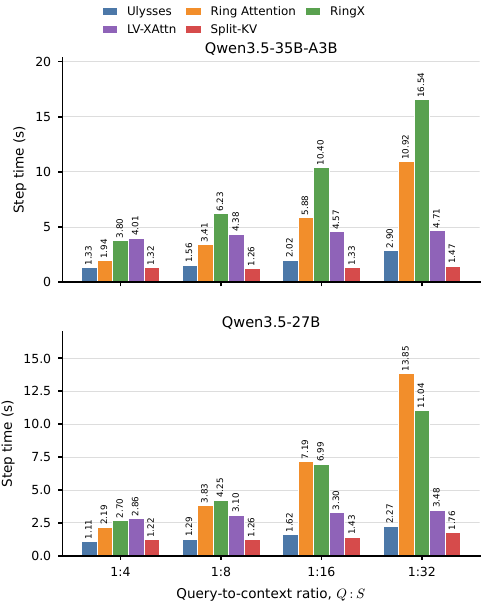}
\caption{Drafter training step time as the query-to-context ratio decreases.
Lower is better; labels report absolute time in seconds.}
\label{fig:qkv-ratio-scaling}
\end{figure}

Figure~\ref{fig:qkv-ratio-scaling} demonstrates the central property of Split-KV: decreasing $Q{:}S$ does not increase its attention communication because only query-sized tensors are exchanged.
Because the figure reports the full training step rather than attention communication alone, residual growth can still come from context-side computation and the larger micro-batch.
For Qwen3.5-35B-A3B, increasing $S/Q$ from 4 to 32 raises Split-KV time by only $1.11\times$, compared with $2.18\times$ for Ulysses, $5.63\times$ for Ring Attention, and $4.35\times$ for RingX; Split-KV is fastest throughout, and its speedup over Ulysses grows from $1.01\times$ to $1.97\times$.
For Qwen3.5-27B, Split-KV grows by $1.44\times$, versus $2.05\times$ for Ulysses, $6.32\times$ for Ring Attention, and $4.09\times$ for RingX.
Ulysses is $1.10\times$ faster at $Q{:}S=1{:}4$, but Split-KV overtakes it at $1{:}8$ and reaches a $1.29\times$ speedup at $1{:}32$.
LV-XAttn grows by only $1.17\times$ and $1.22\times$ for the two models because its communication is also query-sized, but its sequential hop schedule leaves it $1.98\times$--$3.48\times$ slower than Split-KV across all settings.

\subsection{Load-Balancing Quality and Planning Cost}
\label{sec:eval-balance}

We isolate load balancing under a common Split-KV transport, so the strategies differ only in how they assign context chunks.
We fix the per-sample query length at $8\mathrm{k}$ and $b=4$, and evaluate two scaling studies: context-length scaling from $16\mathrm{k}$ to $256\mathrm{k}$ per sample at $P=8$, and CP-size scaling from $P=8$ to 64 at a fixed per-sample context length of $256\mathrm{k}$.
We report percentage imbalance as $100(\rho-1)$, where $\rho=\max_p W_p/\overline{W}$ is the load-imbalance ratio, together with the planning latency required to construct each assignment.
Both drafters use the same anchors, block partition, and query-head count and therefore induce identical work distributions.

\begin{figure}[!t]
\centering
\includegraphics[width=\columnwidth]{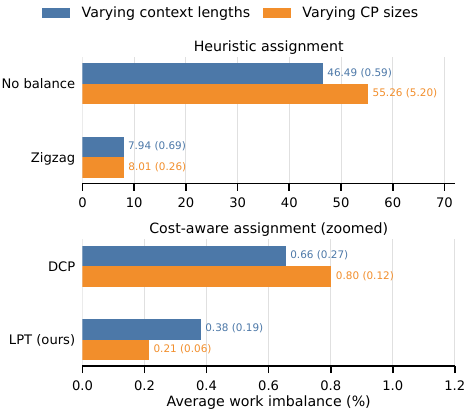}
\caption{Average work imbalance under varying context lengths and CP sizes.
Each label reports $\mathrm{avg}\;(\mathrm{std})$, with both values measured in percentage points.
The lower panel magnifies the cost-aware strategies.}
\label{fig:balance-quality}
\end{figure}

Figure~\ref{fig:balance-quality} summarizes each strategy by its average and standard deviation because their relative ordering is invariant across both scaling studies.
No balance ignores the nonlinear access profile and averages 46.49\% and 55.26\% imbalance under context-length and CP-size scaling.
Zigzag reduces these values to 7.94\% and 8.01\%, but its position-based pairing leaves residual skew.
The cost-aware methods are substantially more accurate: DCP averages 0.66\% and 0.80\%, while LPT achieves the best balance at 0.38\% and 0.21\%.

\begin{figure}[!t]
\centering
\includegraphics[width=\columnwidth]{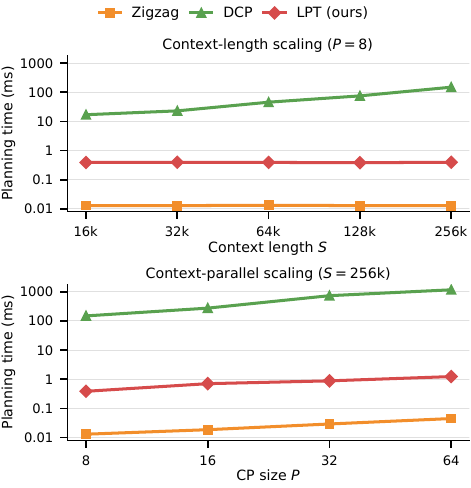}
\caption{Planning latency across context-length and CP-size scaling for Qwen3.5-35B-A3B.
The shared logarithmic axis labels every decade explicitly so that all three planners remain visible.
No balance has zero planning cost and is omitted.}
\label{fig:balance-planning}
\end{figure}

Figure~\ref{fig:balance-planning} shows that comparable assignment quality does not imply comparable planning cost.
For Qwen3.5-35B-A3B, Zigzag takes 0.01--0.05~ms and LPT takes 0.38--1.24~ms across both scaling studies.
DCP instead grows from 16.92 to 148.70~ms with context length and from 148.38 to 1164.64~ms with $P$ because it invokes a general hypergraph partitioner; Qwen3.5-27B exhibits the same trend.

Combining Figures~\ref{fig:balance-quality} and~\ref{fig:balance-planning}, LPT achieves the best load balance with low planning overhead: it is nearly three orders of magnitude faster to plan than DCP while further reducing imbalance.
End to end, LPT achieves a $1.15\times$--$1.42\times$ speedup in drafter training step time over No balance, whose details are extended to Appendix B.

\section{Conclusion}
\label{sec:conclusion}

Diffusion drafter training exposes a distinct CP regime: acceptance length remains stable as context scales with a fixed query budget, while increasing the query budget yields diminishing quality gains, yielding $Q\ll S$; meanwhile, anchor sampling induces nonlinear, sample-dependent attention work.
Split-KV addresses both properties by keeping key/value shards local and limiting per-layer attention communication to query-sized tensors, making this communication independent of context length.
The global query view enables cost-aware LPT assignment that balances the realized workload.
At a CP size of 64, Split-KV retains 85--91\% weak-scaling efficiency and achieves up to $3.67\times$ speedup over the fastest available CP baseline; LPT maintains sub-1\% imbalance with planning latency nearly three orders of magnitude lower than the best available load-balancing baseline, DCP.
These results show that context parallelism for structured attention should jointly design communication and load balancing around the model's training semantics rather than inherit schedules designed for self-attention.

\newpage


\bibliography{main}
\bibliographystyle{mlsys2025}


\end{document}


\twocolumn[
\begin{center}
{\Large\bfseries Split-KV: Efficient Context Parallelism for Training Diffusion Drafters\par}
\vspace{0.5em}
{\large Supplementary Appendix\par}
\end{center}
\vskip 0.2in
]

\appendix

\section{Drafter Quality across Query and Context Lengths}
\label{app:quality-context}

\begin{table*}[!t]
\centering
\caption{Acceptance length on five downstream tasks.
Fix $Q$--Scale $S$ holds 512 anchors per sequence and varies the maximum context length; Fix $S$--Scale $Q$ holds the maximum context length at $64\mathrm{k}$ and varies the anchor budget.
With block size $B=16$, the anchor budgets $\{128,256,512,1024\}$ correspond to query lengths $\{2\mathrm{k},4\mathrm{k},8\mathrm{k},16\mathrm{k}\}$.
Avg. is the unweighted mean of the five displayed task values.}
\label{tab:drafter-context-quality}
\begin{small}
\begin{tabular}{lllrrrrrr}
\toprule
Target model & $S$ & $Q$ & GSM8K & MATH-500 & HumanEval & MBPP & MT-Bench & Avg. \\
\midrule
\multicolumn{9}{l}{\textit{Fix $Q$--Scale $S$}} \\
Qwen3.5-35B-A3B & $4\mathrm{k}$  & $8\mathrm{k}$ & 5.98 & 6.99 & 7.00 & 5.96 & 4.48 & 6.08 \\
 & $16\mathrm{k}$ & $8\mathrm{k}$ & 6.09 & 7.05 & 7.03 & 5.98 & 4.52 & 6.13 \\
 & $64\mathrm{k}$ & $8\mathrm{k}$ & 6.09 & 6.95 & 7.05 & 5.97 & 4.54 & 6.12 \\
Qwen3.5-9B & $4\mathrm{k}$  & $8\mathrm{k}$ & 5.71 & 6.59 & 6.30 & 5.50 & 4.40 & 5.70 \\
 & $16\mathrm{k}$ & $8\mathrm{k}$ & 5.70 & 6.51 & 6.44 & 5.56 & 4.47 & 5.74 \\
 & $64\mathrm{k}$ & $8\mathrm{k}$ & 5.77 & 6.56 & 6.42 & 5.53 & 4.37 & 5.73 \\
\midrule
\multicolumn{9}{l}{\textit{Fix $S$--Scale $Q$}} \\
Qwen3.5-35B-A3B & $64\mathrm{k}$ & $2\mathrm{k}$  & 5.84 & 6.58 & 6.73 & 5.66 & 4.31 & 5.82 \\
 & $64\mathrm{k}$ & $4\mathrm{k}$  & 5.97 & 6.78 & 6.91 & 5.83 & 4.40 & 5.98 \\
 & $64\mathrm{k}$ & $8\mathrm{k}$  & 6.09 & 6.95 & 7.05 & 5.97 & 4.54 & 6.12 \\
 & $64\mathrm{k}$ & $16\mathrm{k}$ & 6.19 & 7.04 & 7.11 & 6.01 & 4.55 & 6.18 \\
Qwen3.5-9B & $64\mathrm{k}$ & $2\mathrm{k}$  & 5.48 & 6.21 & 6.16 & 5.23 & 4.31 & 5.48 \\
 & $64\mathrm{k}$ & $4\mathrm{k}$  & 5.64 & 6.34 & 6.30 & 5.44 & 4.32 & 5.61 \\
 & $64\mathrm{k}$ & $8\mathrm{k}$  & 5.77 & 6.56 & 6.42 & 5.53 & 4.37 & 5.73 \\
 & $64\mathrm{k}$ & $16\mathrm{k}$ & 5.79 & 6.70 & 6.54 & 5.58 & 4.51 & 5.82 \\
\bottomrule
\end{tabular}
\end{small}
\end{table*}

\paragraph{Training setup}
We train DFlash drafters for the frozen Qwen3.5-9B and Qwen3.5-35B-A3B target models on their model-specific regenerated SFT corpora, a mixture of 800K samples from NVIDIA Nemotron Post-Training Dataset V2\footnote{https://huggingface.co/datasets/nvidia/Nemotron-Post-Training-Dataset-v2} and CodeAlpaca\footnote{https://huggingface.co/datasets/HuggingFaceH4/CodeAlpaca\_20K}.
Following the original DFlash recipe \cite{dflash2026}, training uses cross-entropy loss, block size $B=16$, loss-decay factor 7.0, weight decay 0.1, gradient clipping at 1.0, and a cosine learning-rate schedule with a 4\% warmup.
The 35B-A3B drafter uses the original learning rate $6\times10^{-4}$.
For the 9B drafter, this rate causes recurrent gradient spikes, so we reduce it to $2\times10^{-4}$ while leaving the remaining optimization settings unchanged.

Both studies keep the global batch size at 256 and follow the same six-epoch recipe.
\emph{Fix $Q$--Scale $S$} uses 512 anchors per sequence, giving $Q=8\mathrm{k}$ at $B=16$, and varies the maximum per-sequence context length over $\{4\mathrm{k},16\mathrm{k},64\mathrm{k}\}$.
\emph{Fix $S$--Scale $Q$} fixes the maximum per-sequence context length at $64\mathrm{k}$ and varies the anchor budget over $\{128,256,512,1024\}$, giving $Q\in\{2\mathrm{k},4\mathrm{k},8\mathrm{k},16\mathrm{k}\}$.
Each maximum context length is a right-truncation limit rather than a requirement that every sample fill the sequence.
All runs terminate at step 18,684 for Qwen3.5-9B and step 18,744 for Qwen3.5-35B-A3B; the $64\mathrm{k}$, 512-anchor run is shared by the two studies.
All variants use Split-KV for drafter training.

\paragraph{Evaluation protocol}
We evaluate GSM8K, MATH-500, HumanEval, MBPP, and MT-Bench with the official DFlash benchmark and task-specific prompt formatters.
Each drafter--task pair starts a fresh vLLM server on two accelerators with tensor parallelism two, 15 speculative tokens, and at most 32,768 batched tokens.
We issue 128 sequential requests with thinking enabled, at most 2,048 generated tokens, temperature zero, top-$p$ one, and top-$k$ one.
The reported acceptance length equals one plus accepted draft tokens divided by draft steps, including the target model's bonus token.
We measure acceptance length rather than downstream answer accuracy.

\paragraph{Results}
For Fix $Q$--Scale $S$, Table~\ref{tab:drafter-context-quality} shows no monotonic relationship between maximum training context length and acceptance length.
Across $4\mathrm{k}$--$64\mathrm{k}$, the relative task-average span, computed as the difference between the maximum and minimum divided by the minimum, is only 0.82\% for Qwen3.5-35B-A3B and 0.70\% for Qwen3.5-9B, and the best context length varies by task.
Because each variant is evaluated once, we treat these sub-1\% differences as equivalent rather than as statistically significant rankings.
Thus, scaling $S$ does not require scaling $Q$ to preserve drafter quality over the tested range.

For Fix $S$--Scale $Q$, acceptance length increases monotonically with the query budget on both target models, so reducing $Q$ is not free.
However, the marginal gain decreases across successive doublings of $Q$: the mean gains are 0.16, 0.14, and 0.06 for Qwen3.5-35B-A3B and 0.13, 0.12, and 0.09 for Qwen3.5-9B.
Consequently, the final doubling of $Q$ improves mean acceptance length by only 0.98\% and 1.57\%, respectively.
Together, the two studies show that the query budget need not grow with the context and that quality exhibits diminishing returns as that budget increases, establishing a practical $Q\ll S$ regime.

\section{Impact of Load Balancing on Drafter Training Step Time}
\label{sec:eval-balance-e2e}

\begin{figure}[!t]
\centering
\includegraphics[width=\columnwidth]{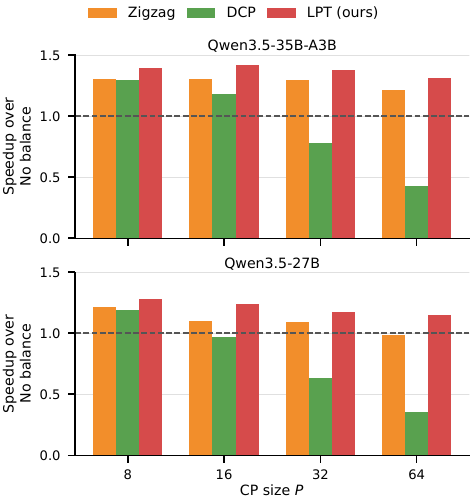}
\caption{Speedup in drafter training step time for each load-balancing strategy over No balance with per-sample context and query lengths of $256\mathrm{k}$ and $8\mathrm{k}$, respectively.
The dashed line denotes performance equal to No balance.}
\label{fig:balance-effect}
\end{figure}

To evaluate the end-to-end effect of load balancing, we fix the per-sample context and query lengths at $256\mathrm{k}$ and $8\mathrm{k}$, respectively, with $b=4$, and measure how assignment quality and planning latency jointly affect drafter training step time as $P$ increases.
Figure~\ref{fig:balance-effect} reports speedup over No balance, so values below one indicate that planning costs more than balancing saves.

LPT is the only strategy that improves both drafters at every scale, yielding $1.15\times$--$1.42\times$ speedup over No balance.
Zigzag remains beneficial except for a slight regression on Qwen3.5-27B at $P=64$, but its residual imbalance prevents it from matching LPT.
DCP performs competitively at small $P$ yet falls below No balance from $P=32$ for Qwen3.5-35B-A3B and from $P=16$ for Qwen3.5-27B as partitioning dominates the shrinking compute time.
At $P=64$, DCP is $3.11\times$ and $3.24\times$ slower than LPT for the two drafters.
Thus, LPT converts near-perfect balance into reductions in drafter training step time without placing an expensive planner on the training critical path.
Although planning can be moved into data preparation for offline drafter training, it remains on the critical path when the drafter is trained online during reinforcement learning \cite{respect2025,li2026breaking}.

\bibliography{main}
\bibliographystyle{mlsys2025}